\documentclass[11pt,a4paper]{article}
\usepackage[hyperref]{emnlp2020}
\usepackage{times}
\usepackage{latexsym}

\usepackage{microtype}

\aclfinalcopy 

\usepackage{amsmath,amsfonts,bm}

\def\eqref#1{equation~\ref{#1}}

\def\1{\bm{1}}

\def\vtheta{{\bm{\theta}}}
\def\valpha{{\bm{\alpha}}}

\def\vc{{\bm{c}}}

\def\ve{{\bm{e}}}
\def\ve{{\bm{e}}}

\def\mA{{\bm{A}}}

\def\mE{{\bm{E}}}

\DeclareMathAlphabet{\mathsfit}{\encodingdefault}{\sfdefault}{m}{sl}
\SetMathAlphabet{\mathsfit}{bold}{\encodingdefault}{\sfdefault}{bx}{n}

\newcommand{\R}{\mathbb{R}}

\usepackage{subcaption}
\usepackage{enumitem}
\usepackage{booktabs}

\usepackage{adjustbox}
\usepackage{multirow}
\usepackage{fontawesome}
\usepackage{nicefrac}
\usepackage{bbold}

\newcommand{\ie}{\textit{i}.\textit{e}., }
\newcommand{\eg}{\textit{e}.\textit{g}., }

\newcommand{\bilm}{\textbb{RAMEN}}
\newcommand{\bilmbase}{\bilm$_{\textsc{base}}$}
\newcommand{\bilmlarge}{\bilm$_{\textsc{large}}$}
\newcommand{\bertbase}{BERT$_{\textsc{base}}$}
\newcommand{\bertlarge}{BERT$_{\textsc{large}}$}
\newcommand{\robertabase}{RoBERTa$_{\textsc{base}}$}
\newcommand{\robertalarge}{RoBERTa$_{\textsc{large}}$}

\usepackage{graphicx}

\newcommand{\fasttext}{fastText}
\newcommand{\fastBPE}{fastBPE}
\newcommand{\fastalign}{fast-align}
\usepackage{xcolor}

\DeclareMathOperator*{\sparsemax}{\mathsf{sparsemax}}
\DeclareMathOperator*{\softmax}{\mathsf{softmax}}

\title{From English to Foreign Languages:\\Transferring Pretrained Language Models}

\author{Ke Tran \\
  Amazon Alexa AI \\
  \texttt{trnke@amazon.com}}

\date{}

\begin{document}
\maketitle
\begin{abstract}
  Pretrained models have demonstrated their effectiveness in many downstream natural language processing (NLP) tasks.
  The availability of multilingual pretrained models enables zero-shot transfer of NLP tasks from high resource languages to low resource ones.
  However, recent research in improving pretrained models focuses heavily on English.
  While it is possible to train the latest neural architectures for other languages from scratch, it is \emph{undesirable} due to the required amount of compute.
  In this work, we tackle the problem of transferring an existing pretrained model from English to other languages under a \emph{limited computational budget}.
  With a single GPU, our approach can obtain a foreign \bertbase{} model within a day (20 hours) and a foreign \bertlarge{} within two days (46 hours).
  Furthermore, evaluating our models on six languages, we demonstrate that our models are better than multilingual BERT on two zero-shot tasks: natural language inference and dependency parsing. Our code is available at \url{https://github.com/anonymized}.
\end{abstract}

\section{Introduction}
\label{sec:intro}
Pretrained models \cite{devlin-etal-2019-bert,peters-etal-2018-deep} have received much of attention recently thanks to their impressive results in many downstream NLP tasks.
Additionally, multilingual pretrained models enable many NLP applications for other languages via zero-short cross-lingual transfer.
Zero-shot cross-lingual transfer has shown promising results for rapidly building applications for low resource languages. Wu and Dredze \shortcite{wu-dredze-2019-beto} show the potential of multilingual BERT \cite{devlin-etal-2019-bert} in zero-shot transfer for a large number of languages from different language families on five NLP tasks, namely, natural language inference, document classification, named entity recognition, part-of-speech tagging, and dependency parsing.

Although multilingual models are an important piece for building up language technology in many languages, recent research on improving pretrained models puts much emphasis on English
\cite{Radford:2019,Dai:2019,yang:xlnet}. The current state of affairs makes it difficult to translate developments in pre-training from English to non-English languages.
To our best knowledge, there are only three publicly available multilingual pretrained models to date:
\begin{enumerate}
  \item The multilingual BERT (mBERT) model \cite{devlin-etal-2019-bert} that supports 104 languages;
  \item Cross-lingual language model (XLM-R)\footnote{\url{https://github.com/facebookresearch/XLM}} \cite{Lample:2019,conneau2019unsupervised} that supports 100 languages;
  \item Language Agnostic SEntence Representations model (LASER)\footnote{\url{https://github.com/facebookresearch/LASER}}  \cite{artetxe:2018x} that supports 93 languages.
\end{enumerate}
Among the three models, LASER is based on neural machine translation approach and strictly requires parallel data to train.

A common practice to train a large-scale multilingual model is to do so from scratch.  \emph{But do multilingual models always need to be trained from scratch? Can we transfer linguistic knowledge learned by English pretrained models to other languages?} In this work, we develop a technique to \textit{rapidly} transfer an existing pretrained model from English to other languages. As the first step, we focus on building a \textit{bilingual} language model (LM) of English and a target language. Starting from a pretrained English LM, we learn the target language specific parameters (\ie word embeddings), while keeping the encoder layers of the pretrained English LM fixed.
We then fine-tune both English and target model to obtain the bilingual LM.
We apply our approach to autoencoding language models with masked language model objective and show the advantage of the proposed approach in zero-shot transfer.
Our main contributions in this work are:
\begin{itemize}
  \item We propose a \emph{fast} adaptation method for obtaining a bilingual \bertbase{} of English and a target language within a day using one Tesla V100 16GB GPU (\S\ref{sec:zero-shot}).
  \item We evaluate our bilingual LMs for six languages on two zero-shot cross-lingual transfer tasks, namely natural language inference (XNL) \cite{conneau-etal-2018-xnli} and universal dependency parsing. We show that our models offer competitive performance or even better that mBERT (\S\ref{sec:results}).
  \item We illustrate that our bilingual LMs can serve as an excellent feature extractor in supervised dependency parsing task (\S\ref{sec:analysis}).
\end{itemize}

Concurrent to our work, \citet{Artetxe:2019onxling} transfer pretrained English model to other languages by fine-tuning only \emph{randomly initialized} target word embeddings while keeping the Transformer encoder fixed. Their approach is simpler than ours but requires more compute (64 TPUv3 chips) to achieve good results.

\section{Bilingual Pretrained LMs}
\label{sec:adaptation}
We first provide some background of pretrained language models.
Let $\mE_e$ be English word-embeddings and $\mathrm{enc}(\vtheta)$ be the Transformer \cite{Vaswani:2017} encoder with parameters $\vtheta$. 
Let $\ve_{w_i}$ denote the embedding of word $w_i$ (\ie $\ve_{w_i} = \mE_e[w_1]$). We omit positional embeddings and bias for clarity.
A pretrained LM typically performs the following computations:
\begin{enumerate}[label=(\roman*)]
  \item transform a sequence of input tokens to contextualized representations
   $[\vc_{w_1},\dots,\vc_{w_n}] = \mathrm{enc}(\ve_{w_1}, \dots, \ve_{w_n}; \vtheta)$
  \item predict an output word $y_i$ at $i^{\text{th}}$ position $p(y_i \,|\, \vc_{w_i}) \propto \exp(\vc_{w_i}^\top \ve_{y_i})$
\end{enumerate}

Autoencoding LM \cite{devlin-etal-2019-bert} corrupts some input tokens $w_i$ by replacing them with a special token \texttt{[MASK]}. It then predicts the original tokens $y_i = w_i$ from the corrupted tokens. Autoregressive LM \cite{Radford:2019} predicts the next token ($y_i = w_{i+1}$) given all the previous tokens. The recently proposed XLNet model \cite{yang:xlnet} is an autoregressive LM that factorizes output with all possible permutations, which shows empirical performance improvement over GPT-2 due to the ability to capture bidirectional context. Here we assume that the encoder performs necessary masking with respect to each training objective.

\begin{figure}[htbp]
  \centering
    \includegraphics[width=\columnwidth]{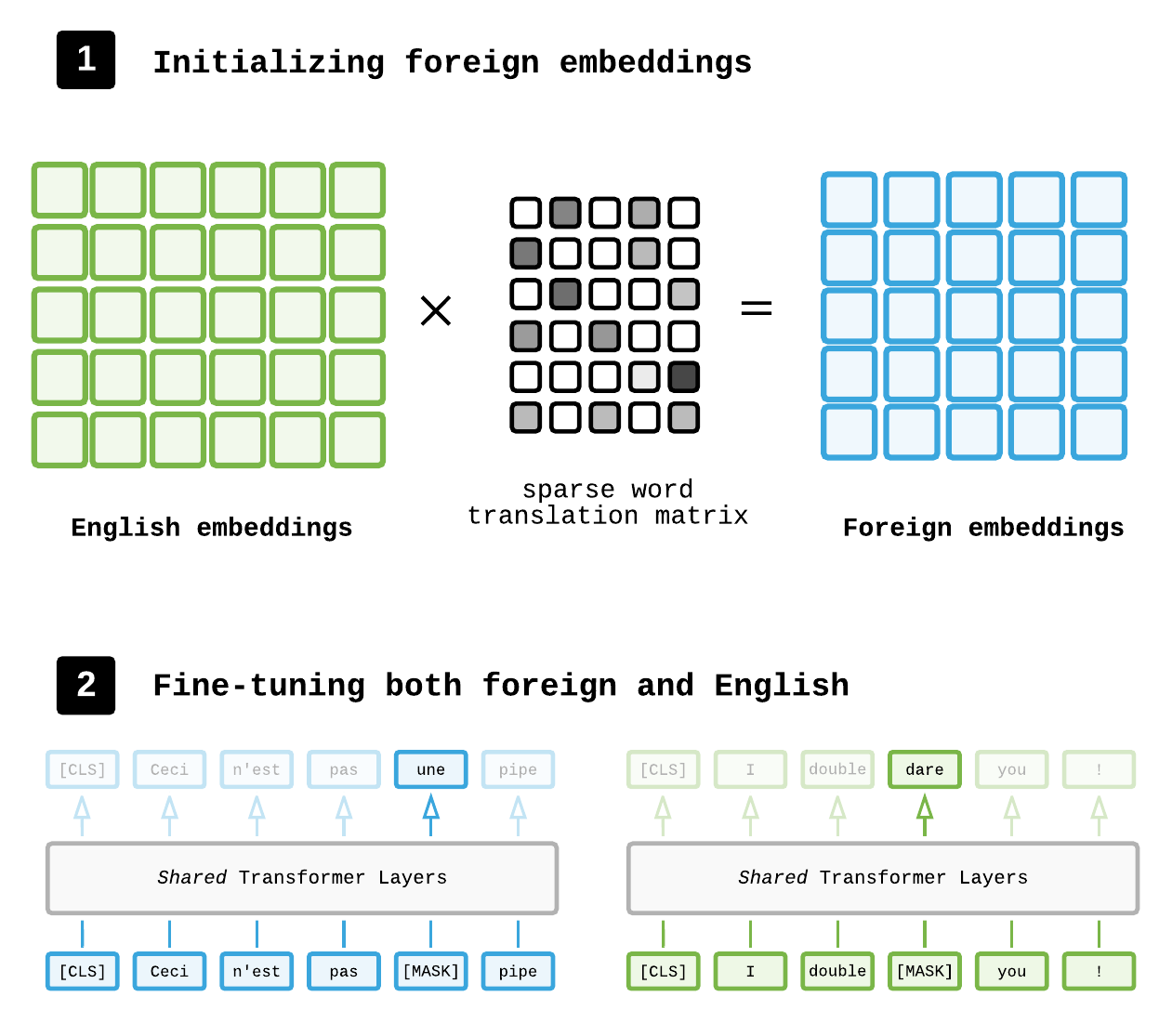}
  \caption{Illustration of our two-step approach. In the first step, foreign embeddings are initialized in English space (\S\ref{ssec:init_tgt_embs}). In the second step, we joinly fine-tune both English and foreign models (\S\ref{ssec:tune_blm}).}
  \label{fig:ramen}
\end{figure}

Given an English pretrained LM, we wish to learn a bilingual LM
for English and a given target language $\ell$ under a limited computational resource budget.
To quickly build a bilingual LM, we directly adapt the English pre-traind model to the target model.
Our approach consists of two steps.
First, we initialize target language word-embeddings $\mE_\ell$ in the English embedding space such that embeddings of a target word and its English equivalents are close together (\S\ref{ssec:init_tgt_embs}).
Next, we construct a bilingual LM of $\mE_e$, $\mE_\ell$, and $\mathrm{enc}(\vtheta)$ and fine-tune all the parameters (\S\ref{ssec:tune_blm}).
Figure~\ref{fig:ramen} illustrates the two steps in our approach.

\subsection{Initializing Target Embeddings}
\label{ssec:init_tgt_embs}
Our approach to learn the initial foreign word embeddings $\mE_\ell \in \R^{|V_\ell| \times d}$ is based on the idea of mapping the trained English word embeddings $\mE_e \in \R^{|V_e| \times d}$ to $\mE_\ell$ such that if a foreign word and an English word are similar in meaning then their embeddings are similar.
We represent each foreign word embedding $\mE_\ell[i] \in \R^d$ as a linear combination of English word embeddings $\mE_e[j] \in \R^d$
\begin{align}
  \mE_\ell[i] &= \sum_{j=1}^{|V_e|} \alpha_{ij} \mE_e[j] = \valpha_i\mE_e\label{eq:universal_rep}
\end{align}
where $\valpha_i\in \R^{|V_e|}$ is a sparse vector and $\sum_j^{|V_e|} \alpha_{ij} = 1$.

In this step of initializing foreign embeddings, having a good estimation of $\valpha$ could speed of the convergence when tuning the foreign model and enable zero-shot transfer (\S\ref{sec:analysis}). In the following, we discuss how to estimate $\valpha_i\;\forall i\in\{1,2, \dots, |V_\ell|\}$ under two scenarios: (i) we have parallel data of English-foreign, and (ii) we only rely on English and foreign monolingual data.

\paragraph{Learning from Parallel Corpus}
Given an English-foreign parallel corpus, we can estimate word translation probability $p(e_j\,|\,\ell_i)$ for any (English-foreign) pair $(e_j, \ell_i)$ using popular word-alignment \cite{brown-etal-1993-mathematics} toolkits such as \fastalign{} \cite{dyer-etal-2013-simple}. We then assign
\begin{align}
  \alpha_{ij} &= p(e_j\,|\, \ell_i)
\end{align}
Since $\valpha_i$ is estimated from word alignment, it is a sparse vector.

\paragraph{Learning from Monolingual Corpus}
For low resource languages, parallel data may not be available. In this case, we rely only on monolingual data (\eg Wikipedias). We estimate word translation probabilities from word embeddings of the two languages. Word vectors of these languages can be learned using \fasttext{}  \cite{bojanowski-etal-2017-enriching} and then are aligned into a shared space with English \cite{lample2018word,joulin-etal-2018-loss}. Unlike learning contextualized representations, learning word vectors is fast and computationally cheap.
Given the aligned vectors $\bar\mE_\ell$ of foreign and $\bar\mE_e$ of English, we calculate the word translation matrix $\mA\in\R^{|V_\ell|\times|V_e|}$ as
\begin{align}
  \mA &= \sparsemax(\bar\mE_\ell \bar\mE_e^\top)\label{eq:sparsemax}
\end{align}
Here, we use $\sparsemax$ \cite{Martins:2016} instead of $\softmax$. $\sparsemax$ is a sparse version of $\softmax$ and it puts zero probabilities on most of the words in the English vocabulary except few English words that are similar to a given foreign word. This property is desirable in our approach since it leads to a better initialization of the foreign word embeddings.

\subsection{Fine-tuning Bilingual LM}
\label{ssec:tune_blm}
We create a bilingual LM by plugging foreign language specific parameters to the pretrained English LM (Figure~\ref{fig:ramen}).
The new model has two separate word embedding layers (and output layers), one for English and one for foreign language.
The encoder layer in between is shared. We then fine-tune this model using English and foreign monolingual data.
Here, we keep tuning the model on English to ensure that it does not forget what it has learned in English and that we can use the resulting model for zero-shot transfer (\S\ref{sec:zero-shot}).
In this step, the encoder is updated so that in can learn syntactic aspects (\ie word order, morphological agreement) of the target languages.

\section{Zero-shot Experiments}
\label{sec:zero-shot}
In the scope of this work, we focus on transferring autoencoding LMs trained with masked language model objective.
We choose BERT and RoBERTa \cite{liu:roberta} as the source models for building our bilingual language models, named \bilm\footnote{The author likes ramen.} for the ease of discussion. We implement our models on top of HuggingFace's Transformer \cite{huggingfaces}.\footnote{\url{https://github.com/huggingface/transformers}}
For each pretrained model, we experiment with 12 layers (\bertbase{} and \robertabase) and 24 layers (\bertlarge{} and \robertalarge) variants.
Using \bertbase{} allows us to compare the results with mBERT model. Using \bertlarge{} and RoBERTa allows us to investigate whether the performance of the target LM correlates with the performance of the source pretrained model.  RoBERTa is a recently published model that is similar to BERT architecturally but with an improved training procedure. By training for longer time, with bigger batches, on more data, and on longer sequences, RoBERTa matched or exceed previously published models including XLNet. We include RoBERTa in our experiments to validate the motivation of our work: \emph{with similar architecture, does a stronger pretrained English model result in a stronger bilingual LM?}
We evaluate our models on two cross-lingual zero-shot tasks: (1) Cross-lingual Natural Language Inference (XNLI) and (2) dependency parsing.

\subsection{Data}
We evaluate our approach for six target languages: French (\texttt{fr}), Russian (\texttt{ru}), Arabic (\texttt{ar}), Chinese (\texttt{zh}), Hindi (\texttt{hi}), and Vietnamese (\texttt{vi}). These languages belong to four different language families. French, Russian, and Hindi are Indo-European languages, similar to English. Arabic, Chinese, and Vietnamese belong to Afro-Asiatic, Sino-Tibetan, and Austro-Asiatic family respectively. The choice of the six languages also reflects different training conditions depending on the amount of monolingual data. French and Russian, and Arabic can be regarded as high resource languages whereas Hindi has far less data and can be considered as low resource.

For experiments that use parallel data to initialize foreign specific parameters, we use the same datasets in the work of Lample and Conneau \shortcite{Lample:2019}. Specifically, we use United Nations Parallel Corpus \cite{Ziemski:16} for \texttt{en-ru}, \texttt{en-ar}, \texttt{en-zh}, and \texttt{en-fr}.
We collect \texttt{en-hi} parallel data from IIT Bombay corpus \cite{kunchukuttan-etal-2018-iit} and \texttt{en-vi} data from OpenSubtitles 2018.\footnote{\url{http://opus.nlpl.eu}}
For experiments that use only monolingual data to initialize foreign parameters, instead of training word-vectors from the scratch, we use the pretrained word vectors\footnote{\url{https://fasttext.cc/docs/en/crawl-vectors.html}} from \fasttext{} \cite{bojanowski-etal-2017-enriching} to estimate word translation probabilities (Eq.~\ref{eq:sparsemax}).
We align these vectors into a common space using orthogonal Procrustes \cite{artetxe-etal-2016-learning,lample2018word,joulin-etal-2018-loss}. We only use \textit{identical words} between the two languages as the supervised signal.
We use WikiExtractor\footnote{\url{https://github.com/attardi/wikiextractor}} to extract extract raw sentences from Wikipedias as monolingual data  for fine-tuning target embeddings and bilingual LMs (\S\ref{ssec:tune_blm}).
We \textit{do not} lowercase or remove accents in our data preprocessing pipeline.

We tokenize English using the provided tokenizer from pretrained models. For target languages, we use \fastBPE\footnote{\url{https://github.com/glample/fastBPE}} to learn 30,000 BPE codes and 50,000 codes when transferring from BERT and RoBERTa respectively.
We truncate the BPE vocabulary of foreign languages to match the size of the English vocabulary in the source models. Precisely, the size of foreign vocabulary is set to 32,000 when transferring from BERT and 50,000 when transferring from RoBERTa.

We use XNLI dataset \cite{conneau-etal-2018-xnli} for classification task and Universal Dependencies v2.4 \cite{ud2.4} for dependency parsing task. Since a language might have more than one treebank in Universal Dependencies, we use the following treebanks: \texttt{en\_ewt} (English), \texttt{fr\_gsd} (French), \texttt{ru\_syntagrus} (Russian), \texttt{zh\_gsd} (Chinese), \texttt{vi\_vtb} (Vietnamese), \texttt{hi\_hdtb} (Hindi), and \texttt{ar\_padt} (Arabic).

\begin{table*}[t]
  \centering
	\scalebox{0.8}{
  \begin{tabular}{@{}l c@{}ccccccccc@{}}
    \toprule
    & \includegraphics[scale=.03]{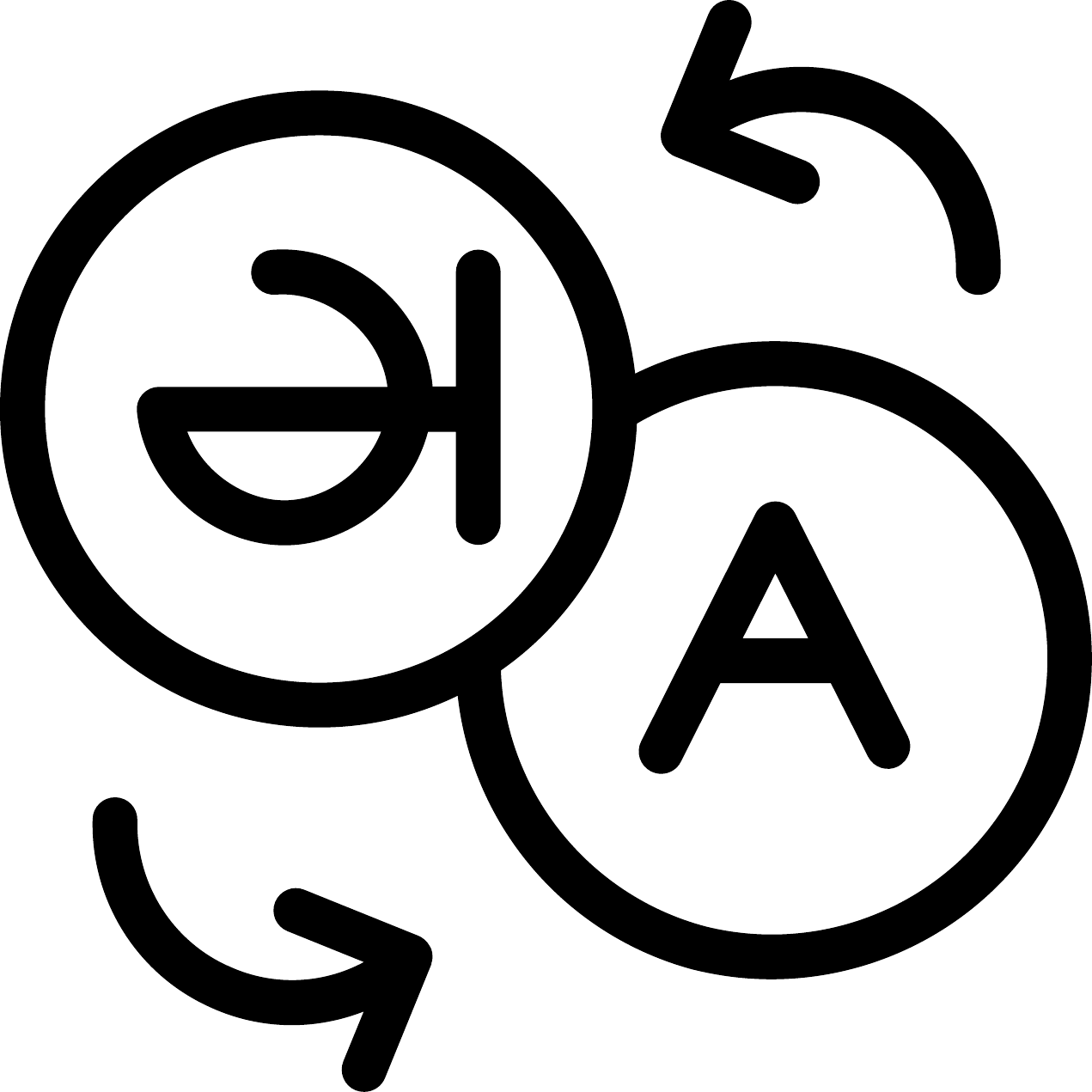} && \texttt{fr} & \texttt{vi} & \texttt{zh}& \texttt{ru} & \texttt{ar} & \texttt{hi} && avg\\
    \midrule
    \cite{conneau-etal-2018-xnli} & \faMinusSquareO     && 67.7 & 66.4 & 65.8& 65.4 & 64.8 & 64.1 && 65.7\\
    \cite{artetxe:2018x} & \faPlusSquareO          && 71.9 & 72.0 & 71.4& 71.5 & 71.4 & 65.5 && 70.6\\
    \cite{Lample:2019} (MLM) &  \faMinusSquareO         && 76.5 & 72.1 & 71.9& 73.1 & 68.5 & 65.7 && 71.3\\
    \cite{Lample:2019} (MLM+TLM) & \faPlusSquareO && 78.7 & 76.1 & 76.5& 75.3 & 73.1 & \bf 69.6 && 74.9\\
    mBERT \cite{wu-dredze-2019-beto} & \faMinusSquareO  && 73.8 & 69.5 & 69.3& 69.0 & 64.9 & 60.0 && 67.8\\
    \midrule
    \bilmbase & & & & & & & & &&\\

    \multirow{2}{*}{\quad + BERT} & \faMinusSquareO  && 75.2 & 71.8 & 70.7 & 71.1 & 69.3 & 62.8 && 70.1\\
    & \faPlusSquareO  && 75.1 & 72.5 & 71.9 & 70.8 & 69.7 & 63.5 && 70.6\\
    \multirow{2}{*}{\quad + RoBERTa} & \faMinusSquareO  && 79.9 & 75.9 & 73.7 & 73.6 & 71.9 & 65.6 && 73.4\\
    & \faPlusSquareO  && 80.3 & 75.6 & 76.2 & \bf 75.8 & 73.1 & 68.1 && 74.9\\
    \addlinespace[.1cm]
    \bilmlarge & & & & && & & \\
    \multirow{2}{*}{\quad + BERT} & \faMinusSquareO  && 78.1 & 74.8 & 74.5 & 73.7 & 70.8 & 64.5 && 72.7\\
    & \faPlusSquareO  && 78.0 & 75.1 & 71.3 & 74.0 & 71.8 & 66.1 && 72.7\\

    \multirow{2}{*}{\quad + RoBERTa} & \faMinusSquareO  && \bf 81.3 & \bf 76.2 & 76.3 & 75.6 & \bf 73.5 & 64.5 && 74.6\\
    & \faPlusSquareO  && 81.0 & \bf 76.2 & \bf 76.8 & 75.0 & 72.9 &  68.2 && 75.0\\
    \bottomrule

  \end{tabular}
	}
  \caption{Zero-shot classification results on XNLI. \faPlusSquareO{} indicates parallel data is used. \bilm{} only uses parallel data for initialization. The best results are marked in \textbf{bold}.}
  \label{tb:xnli}
\end{table*}

\paragraph{Remark on BPE}
\cite{lample2018unsupervised} show that sharing subwords between languages improves alignments between embedding spaces. Wu and Dredze \shortcite{wu-dredze-2019-beto} observe a strong correlation between the percentage of overlapping subwords and mBERT's performances for cross-lingual zero-shot transfer. On the other hand, K {\em et al.}, \shortcite{k2020crosslingual} report in their control experiments that subword overlapping has minimal contribution cross-lingual ability of BERT. Artetxe {\em et al.}, \shortcite{Artetxe:2019onxling} confirms in their study that a shared vocabulary is not necessary for multilingual models. In our current approach, subwords between source and target are \textit{not} shared. A subword that is in both English and foreign vocabulary has two different embeddings.

\subsection{Estimating translation probabilities}
Since pretrained models operate on subword level, we need to estimate subword translation probabilities. Therefore, we subsample 2M sentence pairs from each parallel corpus and tokenize the data into subwords before running fast-align \cite{dyer-etal-2013-simple}.

Estimating subword translation probabilities from aligned word vectors requires an additional processing step since the provided vectors from \fasttext{} are not at subword level.\footnote{In our preliminary experiments, we learned the aligned subword vectors but it results in poor performances.} We use the following approximation to obtain subword vectors:
the vector $\ve_s$ of subword $s$ is the weighted average of all the aligned word vectors $\ve_{w_i}$ that have $s$ as an subword
\begin{align}
  \ve_s &= \sum_{w_j:\, s\in w_j} \frac{p(w_j)}{n_s} \ve_{w_j}
\end{align}
where $p(w_j)$ is the unigram probability of word $w_j$ estimated from monolingual data and $n_s = \sum_{w_j:\, s\in w_j} p(w_j)$.

We take the top 50,000 words in each aligned word-vectors to compute subword vectors.

In both cases, not all the words in the foreign vocabulary can be initialized from the English word-embeddings. Those words are initialized randomly from a Gaussian $\mathcal{N}(0, \nicefrac{1}{d^2})$.

\subsection{Hyper-parameters}
\label{ssec:hyperparams}
In all the experiments, we tune \bilmbase{} for 120,000 updates and \bilmlarge{} for 300,000 updates.
The sequence length is set to 256.
For tuning bilingual LMs, we use a mini-batch size of 112  for \bilmbase{} and 24 for \bilmlarge{} where half of the batch are English sequences and the other half are foreign sequences. This strategy of balancing mini-batch has been used in multilingual neural machine translation \cite{firat-etal-2016-multi,lee-etal-2017-fully}.

We optimize \bilm{} using Adam optimizer. We linearly increase the learning rate from $10^{-7}$ to $10^{-4}$ in the first 4000 updates and then follow an inverse square root decay.
When fine-tuning \bilm{} on XNLI and UD, we use a mini-batch size of 32, Adam's learning rate of $10^{-5}$. The number of epochs are set to 4 and 50 for XNLI and UD tasks respectively.

All experiments are carried out on a single Tesla V100 16GB GPU.
Each \bilmbase{} model is trained within a day and each \bilmlarge{} is trained within two days.\footnote{22 and 46 GPU hours, to be precise. Learning alignment with fast-align takes less than 2 hours and we do not account for training time of \fasttext{} vectors.}

\section{Results}\label{sec:results}
In this section, we present the results of out models for two zero-shot cross lingual transfer tasks: XNLI and universal dependency parsing.

\subsection{Cross-lingual Natural Language Inference}
\label{ssec:eval_xnli}

Table~\ref{tb:xnli} shows the XNLI test accuracy. For reference, we also include the scores from the previous work, notably the state-of-the-art system XLM \cite{Lample:2019}. Before discussing the results, we spell out that the fairest comparison in this experiment is the comparison between mBERT and \bilmbase{}+BERT trained with monolingual only.
We first discuss the transfer results from BERT. Initialized from \fasttext{} vectors, \bilmbase{} slightly outperforms mBERT by 2.6 points on average and widen the gap of 4.4 points on Arabic. \bilmbase{} gains extra 0.3 points on average when initialized from parallel data.
With triple number of parameters, \bilmlarge{} offers an additional boost in term of accuracy and initialization with parallel data consistently  improves the performance.
It has been shown that \bertlarge{} significantly outperforms \bertbase{} on eleven English NLP tasks \cite{devlin-etal-2019-bert}, the strength of \bertlarge{} also shows up when adapted to foreign languages.

Transferring from RoBERTa leads to better zero-shot accuracies. With the same initializing condition, \bilmbase{}+RoBERTa outperforms \bilmbase{}+BERT on average by 3.1 and 4.3 points when initializing from monolingual and parallel data respectively.  This result show that with similar number of parameters, our approach benefits from a better English pretrained model. When transferring from \robertalarge, we obtain state-of-the-art results for five languages. It is worth mentioning that these models are still underfit and they have potential to push the XNLI performance further.

Currently, \bilm{} only uses parallel data to initialize foreign embeddings.
\bilm{} can also exploit parallel data through translation objective proposed in XLM. We believe that by utilizing parallel data during the fine-tuning of \bilm{} would bring additional benefits for zero-shot tasks. We leave this exploration to future work.
In summary, starting from \bertbase{}, our approach obtains competitive bilingual LMs with mBERT for zero-shot XNLI. Our approach shows the accuracy gains when adapting from a better pretrained model.

\subsection{Universal Dependency Parsing}
\label{ssec:eval_ud}
We build on top of \bilm{} a graph-based dependency parser \cite{Dozat:2016}. For the purpose of evaluating the contextual representations learned by our model, we \textit{do not} use part-of-speech tags. Contextualized representations are directly fed into Deep-Biaffine layers to predict arc and label scores. Table~\ref{tb:ud} presents the Labeled Attachment Scores (LAS) for zero-shot dependency parsing.

\begin{table}[ht]
  \centering
	\scalebox{0.7}{
  \begin{tabular}{@{}lc@{}ccccccccc@{}}
    \toprule
    & \includegraphics[scale=.03]{figures/bitexta} && \texttt{fr} & \texttt{vi} & \texttt{zh}& \texttt{ru} & \texttt{ar} & \texttt{hi} && avg\\
    \midrule
    mBERT & \faMinusSquareO  &&   71.6 & 35.7 & 26.6 &  65.2 &  36.4 & 30.4 && 44.3\\
    \midrule
    \bilmbase & & & & & & & & &&\\

    \multirow{2}{*}{\quad + BERT} & \faMinusSquareO  && 78.0 & 38.3 & 30.1 & 67.2 & 40.6 & 38.6 && 48.8\\
    & \faPlusSquareO  && 77.0 & 36.7 & 30.8 & 66.8 & 40.9 & 40.9 && 48.9 \\

    \multirow{2}{*}{\quad + RoBERTa} & \faMinusSquareO  && 79.1 & 38.9 & 31.5 & \bf 67.7 & 42.2 & 41.2 && 50.1\\
    & \faPlusSquareO  && 78.5 & 39.4 & 31.3 & 65.1 & 41.4 & 44.2 &&  50.0\\

    \addlinespace[.1cm]
    \bilmlarge & & & & & & & & &&\\

    \multirow{2}{*}{\quad + BERT} & \faMinusSquareO && 78.2 & 39.5 & 30.5 & 65.6 & 45.3 & 44.0 && 50.5\\
    & \faPlusSquareO  && 78.9 & 38.4 & 30.5 & 66.3 & 43.4 & 44.1 && 50.3\\

    \multirow{2}{*}{\quad + RoBERTa} & \faMinusSquareO  && \bf 79.7 & \bf 40.0 & \bf 32.2 & 65.7 & \bf 44.1 & 44.6 && 51.1\\
    & \faPlusSquareO  && 79.1 & 39.2 & 30.5 & 65.4 & 43.8 & \bf 46.8 && 50.8\\
    \bottomrule
  \end{tabular}
	}
  \caption{LAS scores for zero-shot dependency parsing. \faPlusSquareO{} indicates parallel data is used for initialization. Punctuation are removed during the evaluation. The best results are marked in \textbf{bold}.}
  \label{tb:ud}
\end{table}

We first look at the fairest comparison between mBERT and monolingually initialized \bilmbase{}+BERT. The latter outperforms the former on all the languages with the average of 4.5 LAS point. We observe the largest gain of +8.2 for Hindi and +6.2 LAS for French. Arabic enjoys +4.2 LAS from our approach.
With similar architecture (12 or 24 layers) and initialization (using monolingual or parallel data), \bilm{}+RoBERTa performs better than \bilm{}+BERT for most of the languages. Arabic and Hindi benefit the most from bigger models. For the other four languages, \bilmlarge{} renders a modest improvement over \bilmbase.

\section{Analysis}
\label{sec:analysis}
Throughout this section, we carry out the main analysis for the \bilmbase{}+BERT model that only use monolingual data.

\subsection{How does linguistic knowledge transfer happen through each training stages?}
\label{ssec:language_acquisition}
\begin{figure}[h]
    \centering
    \begin{subfigure}[b]{0.35\textwidth}
        \includegraphics[width=\textwidth]{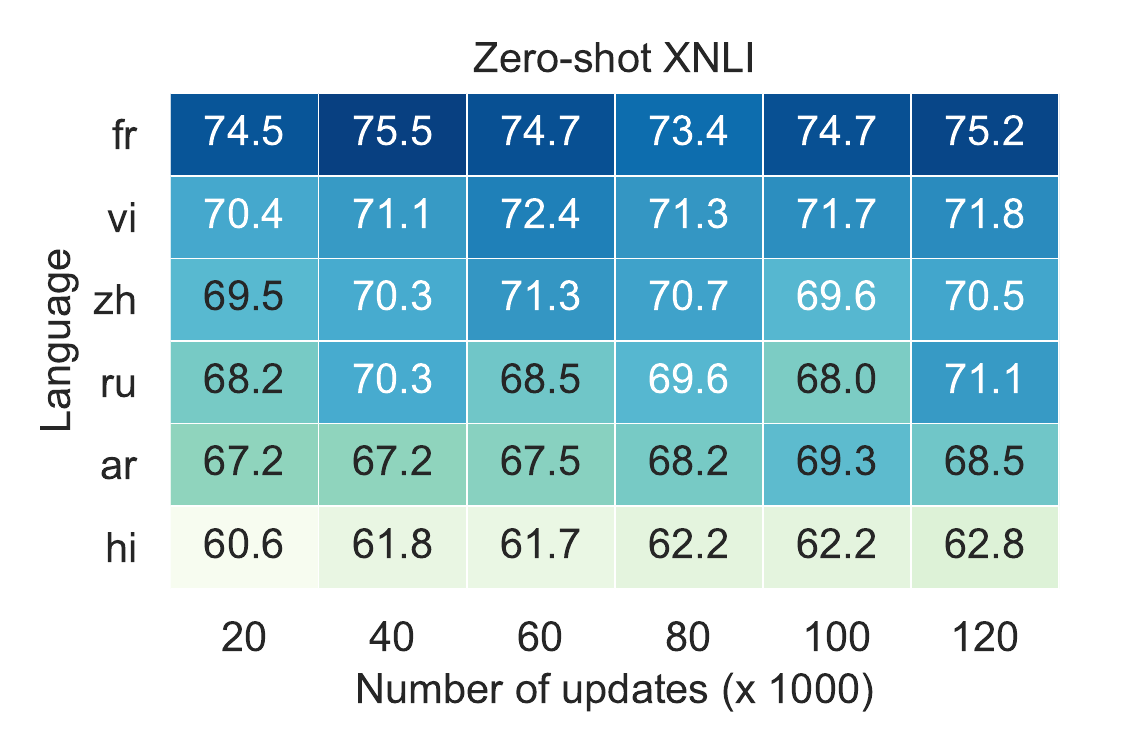}
    \end{subfigure}
    \quad
    \begin{subfigure}[b]{0.35\textwidth}
        \includegraphics[width=\textwidth]{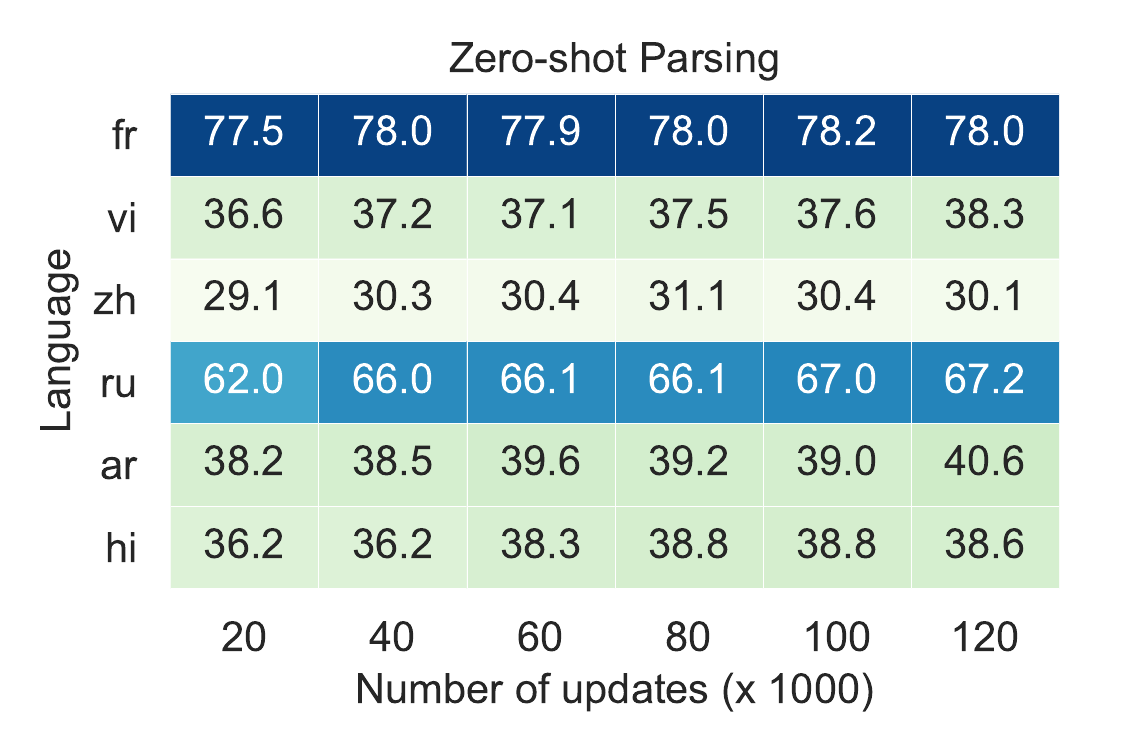}
    \end{subfigure}
    \caption{Accuracy and LAS evaluated at each checkpoints.}
    \label{fig:perf}
\end{figure}

We evaluate the performance of \bilm{}+\bertbase{} (initialized from monolingual data) at each 20K training steps. The results are presented in Figure~\ref{fig:perf}.

We observe the general trend that futher fine-tuning \bilm{} improves the performances of two zero-shot tasks.  Just after 40,000 fine-tuning updates\textemdash corresponding to seven training hours\textemdash \bilmbase{}+BERT surpasses mBERT on both XNLI (Table~\ref{tb:xnli}) and universal dependency parsing (Table~\ref{tb:ud}). Interstingly, we find that the English encoder can quickly adapt to the foreign syntax that has different word order (\texttt{ar} and \texttt{hi}) with a little training data.

Language similarities seem to have more impact on transferring syntax than semantics.
French enjoys 78.0 LAS for being closely related to English, whereas Arabic and Hindi, SOV languages, modestly reach 38.2 and 36.2 points using the SVO encoder. Although Chinese has SVO order, it is often seen as head-final while English is strong head-initial. Perhaps, this explains the poor performance for Chinese.

\subsection{Impact of initialization}
\label{ssec:init_impact}
Initializing foreign embeddings is the backbone of our approach. A good initialization leads to better zero-shot transfer results and enables fast adaptation.
To verify the importance of a good initialization, we train a \bilmbase{}+BERT with foreign word-embeddings that are initialized randomly from $\mathcal{N}(0, \nicefrac{1}{d^2})$. For a fair comparison, we use the same hyper-parameters in \S\ref{ssec:hyperparams}. Table~\ref{tb:init} shows the results of XNLI and UD parsing of random initialization. In comparison to the initialization using aligned \fasttext{} vectors, random initialization decreases the zero-shot performance of \bilmbase{} by 10.3\% for XNLI and 11.6 points for UD parsing on average. We also see that zero-shot parsing of SOV languages (Arabic and Hindi) suffers random initialization.
\begin{table}[h!]
  \centering
  \scalebox{0.7}{
  \begin{tabular}{@{}lccccccccccc@{}}\toprule
     & \includegraphics[scale=.2]{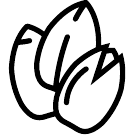} && \texttt{fr} & \texttt{vi} & \texttt{zh} & \texttt{ru} & \texttt{ar} & \texttt{hi} && avg\\
    \midrule
    \multirow{3}{*}{XNLI}
    &\texttt{rnd} &&  66.9 & 67.1 &  65.7 & 59.1 & 50.8 & 48.5 && 59.7\\
    &\texttt{scr} &&  71.1 & 58.0 &  62.8 & 63.8 & 62.1 & 47.5 && 60.8\\
    &\texttt{vec} &&  75.2 & 71.8 &  70.5 & 71.1 & 68.5 & 62.8 && 70.0\\
    \midrule
    \multirow{3}{*}{UD} &\texttt{rnd} &&  71.0 & 33.2 &  25.5 & 60.3 & 19.7 & 13.4 && 37.2\\
    &\texttt{scr} &&  63.8 & 17.0 &  13.6 & 61.0 & 15.5 & 11.2 && 30.3\\
    &\texttt{vec} &&  78.0 & 38.3 &  30.1 & 67.2 & 40.6 & 38.6 && 48.8\\
    \bottomrule
  \end{tabular}
  }
  \caption{Comparison between random initialization (\texttt{rnd}) of language specific parameters and initialization using aligned \fasttext{} vectors (\texttt{vec}) and bilingual BERT trained from scratch (\texttt{scr}) for 400 hours.}
  \label{tb:init}
\end{table}

To highlight the efficiency of our transferring approach, we compare \bilmbase+BERT with a bilingual BERT (bBERT) trained from scratch (\texttt{scr}).
For a fair comparision, we use the hyperparameters described in section \S\ref{ssec:hyperparams}. All bBERT models are trained for 2,300,000 updates (400 GPU hours), which is more than sixteen times longer than RAMEN. We emphasize two important observations from Table~\ref{tb:init}. First, the randomly initialized foreign embeddings \bilm{} performs on par with bBERT on XNLI task and significantly better than bBERT on UD parsing. This suggests that a large amount of linguistic knowledge can be transfered through reusing the pretrained BERT encoder. Secondly, with a careful initialization of foreign embeddings, our \bilmbase+BERT  outshines bBERT with even just 20,000 updates (3.5 GPU hours) as shown in Figure~\ref{fig:perf}.

\subsection{Are contextual representations from \bilm{} also good for supervised parsing?}
\label{ssec:sup_eval}

All the \bilm{} models are built from English and tuned on English for zero-shot cross-lingual tasks. It is reasonable to expect \bilm s do well in those tasks as we have shown in our experiments. \textit{But are they also a good feature extractor for supervised tasks?} We offer a partial answer to this question by evaluating our model for supervised dependency parsing on UD datasets.

\begin{table}[ht!]
  \centering
  \scalebox{0.7}{
  \begin{tabular}{@{}l c c c c c c c cc@{}}\toprule
    && \texttt{fr}  & \texttt{vi} & \texttt{zh} & \texttt{ru} & \texttt{ar} & \texttt{hi} && avg\\
    \midrule
    mBERT   && 92.1 & 62.2 &  85.1 &  93.1 & 83.6 & 91.3 && 84.6 \\
    \midrule
    \bilmbase & & & & & & \\
    \quad + BERT && 92.2 &  63.3 & 85.0 &  93.3 & 83.9 &  92.2 && 85.0 \\
		\quad + RoBERTa && 92.8 &  65.3 & 86.0 &  93.7 &  84.9 &  92.5 && 85.9\\

		\addlinespace[.1cm]
    \bilmlarge & & & & & & &  \\
    \quad + BERT && 92.8 & 64.7 & 86.2 & 93.9 & 84.9 & 92.0 &&  85.7 \\
		\quad + RoBERTa && 93.0 & 66.4 & 87.3 & 94.0 & 85.3 & 92.8 && 86.5\\
    \bottomrule
  \end{tabular}
  }
  \caption{Evaluation in supervised UD parsing. The scores are LAS.}
  \label{tb:ud_supervised}
\end{table}

We used train/dev/test splits provided in UD to train and evaluate our \bilm-based parser.
Table~\ref{tb:ud_supervised} summarizes the results (LAS) of our supervised parser. For a fair comparison, we choose mBERT as the baseline and all the \bilm{} models are initialized from aligned \fasttext{} vectors. With the same architecture of 12 Transformer layers, \bilmbase{}+BERT performs competitive to mBERT and outshines mBERT by +1.1 points for Vietnamese. The best LAS results are obtained by \bilmlarge{}+RoBERTa with 24 Transformer layers. Overall, our results indicate the potential of using contextual representations from \bilm{} for supervised tasks.

\section{Conclusions}
In this work, we have presented a simple and effective approach for rapidly building a bilingual LM under a limited computational budget. Using BERT as the starting point, we demonstrate that our approach performs better than mBERT on two cross-lingual zero-shot sentence classification and dependency parsing. We find that the performance of our bilingual LM, \bilm, correlates with the performance of the original pretrained English models. We also find that \bilm{} is also a powerful feature extractor in supervised dependency parsing. Finally, we hope that our work sparks of interest in  developing fast and effective methods for transferring pretrained English models to other languages.

\bibliography{emnlp2020}
\bibliographystyle{acl_natbib}

\end{document}